\def\year{2020}\relax
\documentclass[letterpaper]{article} % DO NOT CHANGE THIS
\usepackage{aaai20}  % DO NOT CHANGE THIS
\usepackage{times}  % DO NOT CHANGE THIS
\usepackage{helvet} % DO NOT CHANGE THIS
\usepackage{courier}  % DO NOT CHANGE THIS
\usepackage[hyphens]{url}  % DO NOT CHANGE THIS
\usepackage{graphicx} % DO NOT CHANGE THIS
\usepackage{graphicx}  % DO NOT CHANGE THIS
\usepackage{float}

\usepackage{booktabs}
\usepackage{makecell}
\usepackage{tabularx}
\newcommand{\ra}[1]{\renewcommand{\arraystretch}{#1}}

\newenvironment{changemargin}[2]{%
\begin{list}{}{%
\setlength{\topsep}{0pt}%
\setlength{\leftmargin}{#1}%
\setlength{\rightmargin}{#2}%
\setlength{\listparindent}{\parindent}%
\setlength{\itemindent}{\parindent}%
\setlength{\parsep}{\parskip}%
}%
\item[]}{\end{list}}

\usepackage{titlesec}
\usepackage{amsmath}
\usepackage{amssymb}

\newcommand{\citet}[1]{\citeauthor{#1} \shortcite{#1}}
\newcommand{\citep}{\cite}

\usepackage{booktabs}
\usepackage{makecell}
\usepackage{tabularx}

\title{Hierarchical Reinforcement Learning for Open-Domain Dialog}
\author{Abdelrhman Saleh,\textsuperscript{\rm 1,}\thanks{Equal Contribution} Natasha Jaques,\textsuperscript{\rm 2,}\textsuperscript{\textasteriskcentered} Asma Ghandeharioun,\textsuperscript{\rm 2} \\
\Large \textbf{Judy Hanwen Shen,\textsuperscript{\rm 2} Rosalind Picard\textsuperscript{\rm 2}} \\ 
\textsuperscript{\rm 1}Harvard University, \textsuperscript{\rm 2}MIT Media Lab\\
abdelrhman\_saleh@college.harvard.edu, jaquesn@mit.edu, \\ % email address 
asma\_gh@mit.edu, judyshen@mit.edu, picard@media.mit.edu
}
 
\begin{document}

\maketitle

\begin{abstract}
Open-domain dialog generation is a challenging problem; maximum likelihood training can lead to repetitive outputs, models have difficulty tracking long-term conversational goals, and training on standard movie or online datasets may lead to the generation of inappropriate, biased, or offensive text. Reinforcement Learning (RL) is a powerful framework that could potentially address these issues, for example by allowing a dialog model to optimize for reducing toxicity and repetitiveness. However, previous approaches which apply RL to open-domain dialog generation do so at the word level, making it difficult for the model to learn proper credit assignment for long-term conversational rewards. In this paper, we propose a novel approach to hierarchical reinforcement learning (HRL), VHRL, which uses policy gradients to tune the utterance-level embedding of a variational sequence model. This hierarchical approach provides greater flexibility for learning long-term, conversational rewards. We use self-play and RL to optimize for a set of human-centered conversation metrics, and show that our approach provides significant improvements -- in terms of both human evaluation and automatic metrics -- over state-of-the-art dialog models, including Transformers.  
\end{abstract}

\section{Introduction}
\label{sec:intro}
Since the inception of the Turing test, generating convincing open-ended (open-domain) dialog has been a fundamental challenge in artificial intelligence (AI).  %Solving this problem requires more than an effective language model; it requires improved understanding of human conventions, affect, and the dyadic nature of a good conversation. Yet 
A successful open-domain dialog system could provide enormous value %, not only as a hallmark of AI, but also 
enabling more natural human-computer interaction. Successful dialog models could also unlock new, beneficial applications of AI, such as companion chatbots for therapy applications.
%designed to improve mental health. %(e.g. \citet{fitzpatrick2017delivering}).
%Open-domain dialog models attempt to learn a generative model of conversations, potentially allowing them to produce interesting and varied

However, current generative models for dialog suffer from several shortcomings that limit their usefulness in the real world. Training on standard dialog datasets collected online or from movie scripts often leads to malicious, aggressive, biased, or offensive responses \cite{curry2018metoo,he2018detecting,henderson2018ethical,wallace2019universal}. There are no guarantees on the quality and  sensitivity of the generated text, often preventing open-domain dialog systems from being deployed for safety-critical applications such as mental health. Further, maximum likelihood estimation (MLE) training of such models often leads to the generation of dull and repetitive text \cite{li2016deep}. Finally, models may have difficulty tracking long-term aspects of the conversation, and evidence has shown that they do not adequately condition on the conversation history in generating responses \cite{sankar2019neural}.

Reinforcement Learning (RL) is a powerful paradigm that allows dialog models to optimize for non-differentiable metrics of conversation quality, and thereby helps overcome the above problems. In this paper, we use RL to learn from self-play; the model talks to a fixed copy of itself, and computes reward functions on the generated conversation. We propose novel, human-centered rewards, such as minimizing \textit{toxicity} of a conversation, %\footnote{Toxicity is estimated by training a model on the data available at \url{https://www.kaggle.com/c/jigsaw-toxic-comment-classification-challenge}}, 
in order to %guide models towards safe regions of the training data distribution and 
limit inappropriate responses. We also design rewards based on the psychology of good conversation (e.g. \citet{bodie2012listening,weger2010active}), %bodie2015role
and reward recently proposed conversation metrics that are associated with improved human judgments of conversation quality \cite{see2019makes}. Table \ref{tab:samples} shows how RL can be used to optimize our proposed metrics, shifting the conversation towards more positive and engaging conversation, with less repetition and toxicity.

\begin{table*}\centering
% \ra{0.7}
\noindent\makebox[\textwidth]{
\begin{tabular}{p{2.55cm}|p{2.45cm}|p{2.4cm}|p{2.52cm}|p{2.45cm}|p{2.25cm}}
\normalsize Transformer & Sentiment & Question & Repetition & Semantic Sim.& Toxicity \\ 
\midrule

\small
\makecell[lc]{[Usr]: hello!
\\ {[}Bot{]}: i don't know.
\\ {[}Usr{]}: how are you?
\\ {[}Bot{]}: i don't know \\what that means.} & 

\small
\makecell[lc]{[Usr]: hello!
\\ {[}Bot{]}: i'm in my \\sunroom a lot and \\it's a great feeling!
\\ {[}Usr{]}: sounds fun
\\ {[}Bot{]}: i'm excited!} & 

\small
\makecell[lc]{[Usr]: hi there!
\\ {[}Bot{]}: what do \\you do for fun?\\
{[}Usr{]}: i read books \\
{[}Bot{]}: what do you \\like to do in your \\life?}& 

\small
\makecell[lc]{[Usr]: how are you?
\\ {[}Bot{]}: i'm a \\ reservoir engineer
\\ {[}Usr{]}: cool. i am \\still in college 
\\ {[}Bot{]}: today is my \\first day of vacation
}&

\small
\makecell[lc]{[Usr]: i like to\\ listen to ed sheeran
\\ {[}Bot{]}: and what \\genre is it?
\\{[}Usr{]}: pop
\\ {[}Bot{]}: any \\recommendations?} & 

\small
\makecell[lc]{[Usr]: you're so \\stupid
\\ {[}Bot{]}: you're cute
\\ {[}Usr{]}: i hate you
\\ {[}Bot{]}: but i don't \\know why} \\
\bottomrule
\end{tabular}}
\caption{Chat examples showing that our proposed VHRL technique can be used to optimize for human-centered rewards.}
\label{tab:samples}
\end{table*}

Applying RL to open-domain dialog generation is a challenging problem. Most prior approaches (e.g. \citet{li2016deep,li2017adversarial,jaques2019way,ranzato2015sequence,yu2017seqgan}) learn to model rewards at the word level, meaning that the reward is applied to affect the probability of generating each word in the response. Such low-level control makes credit assignment especially challenging, since high-level rewards based on multiple conversation turns must be applied to specific words. Further, this approach makes the number of actions available to the RL agent equivalent to the number of words in the dialog model's vocabulary (20,000 in our case). Thus there are several orders of magnitude more actions than in typical applications of deep RL, such as learning to play an Atari game with only tens of actions (e.g. \citet{kulkarni2016hierarchical}). %mnih2013playing 

To overcome these challenges, we leverage hierarchical reinforcement learning (HRL) to model rewards at the utterance level, improving the flexibility of dialog models to learn long-term, conversational rewards. Specifically, we propose a novel approach, Variational Hierarchical Reinforcement Learning (VHRL), which uses policy gradients to adjust the prior probability distribution of the latent variable learned at the utterance level of a hierarchical variational model. We show that this approach allows for improved learning of conversational rewards that are not modeled well at the word level.

To evaluate our models, we not only compute automatic metrics, but conduct an interactive human evaluation using the \url{https://neural.chat/} platform \cite{ghandeharioun2019approximating}, in which humans chat live with our bots about anything they choose. This represents a more realistic test of real-world generalization performance than is typically employed when testing RL models in the same environment in which they were trained. Our evaluation reveals that VHRL improves human judgments of conversational quality above state-of-the-art dialog architectures, including Transformer-based models. 

In summary, the paper makes the following contributions: a) Develops a new technique, VHRL, for hierarchical control of variational dialog models; b) Demonstrates the effectiveness of training open-domain dialog models with VHRL and self-play, showing improvements over state-of-the-art dialog architectures with both human evaluation and automatic metrics; and c) %Introduces reducing toxicity as a reward function that can ensure conversations are more appropriate and safe for real-world deployment.
Introduces and compares several reward functions for guiding conversations to be less toxic and repetitive, and more engaging, positive, contingent on user input.
% \begin{itemize}
%     \item Develops a new technique, VHRL, for hierarchical control of variational dialog models.%for improved modeling of long-term rewards. 
%     \item Demonstrates the effectiveness of training open-domain dialog models with VHRL and self-play, showing improvements over state-of-the-art dialog architectures with both human evaluation and automatic analysis. 
%     \item Introduces reducing toxicity as a reward function that can ensure conversations are more appropriate and safe for real-world deployment.
% \end{itemize}
In addition, we release code for our evaluation platform and our models at \url{https://github.com/natashamjaques/neural_chat}.

\section{Related Work}
Open-domain dialog systems currently lack reliable automatic evaluation metrics \cite{liu2016not,lowe2017towards}. Recently, authors have begun to propose new metrics of conversation quality (e.g. \citet{see2019makes,hancock2019learning,zhou2018design}), and have even proposed evaluating metrics on conversations generated with self-play \cite{ghandeharioun2019approximating}. However, these studies have not attempted to directly optimize for their proposed conversation metrics with RL.

There has been significant progress in improving dialog generation outside of RL. One popular approach for controllable generation is conditional training, where a learned embedding vector is associated with a desired output feature and fed into the decoder to control generation (e.g. \citet{see2019makes,colombo2019affect,huang2018automatic}). %ko2019linguistically
This approach has multiple limitations. First, the model can only learn associations present in the training data, and cannot explore to discover improved methods for optimizing the desired features. Second, the conditional embeddings are learned at training time with teacher forcing and thus suffer from exposure bias \cite{ranzato2015sequence}. Using RL avoids these limitations as it allows exploring regions of space not present in the training data and directly optimizing for rewards at inference time. %which is more indicative of the model's real world behavior. 
In addition, RL learns the total expected future reward of taking some action now, given how the rest of the conversation will unfold in the future. This allows RL models to make long-term trade-offs about the benefits of generating words and utterances in the conversation context. Finally, our approach does not require the addition of any new parameters or complex components. Instead, it can be use to tune pre-existing models to output better, more appropriate responses.

\subsection{Reinforcement Learning for Dialog}
Improving dialog models with RL is a difficult problem, and past work has largely been restricted to task-oriented dialog systems, which have a limited number of task-specific actions (e.g. 
%\citet{gavsic2011line}, 
\citet{liu2017iterative,su2017sample}). %liu2018dialogue
Attempts to apply RL to open-domain dialog generation are less common. Even in this setting, authors may choose to use a highly restricted action space, for example, using RL to choose dialog acts for conditional generation \cite{xu2018towards}. 

\citet{li2016deep} applied deep RL to the full vocabulary-sized action space, optimizing for rewards such as \textit{ease of answering}. RL has also been used to optimize for rewards based on GAN-like discriminators trained to distinguish human-generated from model-generated text 
\cite{li2017adversarial,yu2017seqgan}. %li2018dialogue 
Note that it is difficult to apply the traditional GAN approach of backpropagating discriminator gradients directly into the generator, because the sampling procedure used to generate each word does not preserve gradients. This makes RL an attractive option for improving dialog generation since it can optimize for non-differentiable rewards. 

Sentiment has been used as a reward in an RL setting for dialog \citep{shin2019happybot}. \citet{jaques2019way} optimize for sentiment and several other conversation metrics by learning from a static batch of human-bot conversations using Batch RL. We believe we are the first to propose using RL to reduce \textit{toxicity} in an open-domain dialog setting, in order to ensure the model produces more appropriate and safe conversations.

Hierarchical models have been investigated extensively in the context of language modeling.  %\cite{serban2017hierarchical,shen2018improving,zhao2017learning}. 
These models take advantage of the natural hierarchical structure of language, decomposing input into utterances at one level, and words at another. However, attempts to apply hierarchical RL (HRL) to dialog generation have so far been limited to task-oriented dialog systems \cite{zhang2018multimodal,peng2017composite,budzianowski2017sub,tang2018subgoal}. % could cut zhang
To the best of our knowledge, we are the first to apply HRL to open-domain dialog generation.

% An advantage of applying RL to dialog is that we directly optimize for desired rewards at the inference phase with the model's outputs conditioned on its previous predictions in an autoregressive fashion. This avoids the exposure bias of teacher forcing in maximum likelihood training and is more representative of test time behavior. 

\subsection{Hierarchical Reinforcement Learning}
Many approaches have been proposed for building hierarchical agents within the context of reinforcement learning for games and robotics \cite{sutton1999between,precup2001temporal,dietterich2000hierarchical,vezhnevets2017feudal,bacon2017option,nachum2018data}. %dayan1993feudal 
The options framework proposed by \citet{sutton1999between} is one popular approach for HRL. 
% First, at the top level, a manager \textit{policy over options}  (or manager) selects a 
At the bottom level of the hierarchy, a set of options (or workers) which are \textit{policies over actions} interact with the environment until terminated by the agent. At the top level, a \textit{policy over options} (or manager) selects options to be executed until termination, at which point another option is picked and the process is repeated. 
% HRL agents make decisions at two levels of temporal abstraction as the workers interact with the environment at every time step while the manager waits until a worker terminates to pick another 
The different levels of temporal abstraction introduced by this hierarchy allows for better long-term planning relative to traditional, flat reinforcement learning techniques. 

A major focus of HRL has been on sub-goal or option discovery for training worker policies. Bottom-level policies are often learned using handcrafted sub-goals \cite{kulkarni2016hierarchical,tessler2017deep}, intrinsic rewards \cite{vezhnevets2017feudal}, or pseudo-rewards \cite{dietterich2000hierarchical}, while the manager policy is learned using extrinsic rewards from the environment. Our approach also allows for optimizing different rewards at different levels of the hierarchy, thus creating distinct goals for the worker and the manager. However, unlike other HRL approaches we expose both the worker and manager policies to extrinsic rewards and add weight hyper-parameters to regulate the effect of the rewards at each level. This remedies a weakness of pseudo-reward methods where a worker only focuses on achieving its sub-goals while disregarding the effect on the extrinsic environment reward.

% \citet{vezhnevets2017feudal} train a manager to identify directional goals and intrinsically reward a worker for shifting the environment's state in these directions. Similarly, \citet{nachum2018data} use the environment's states directly as goals and the worker's sub-goal is to reach the selected states. These approaches achieve impressive results on physical navigation-based tasks such as collecting keys in Montezuma's Revenge. The goals defined in our approach do not carry the same semantic meaning since the physical grounding of goals does not apply in dialog. Although our approach allows for optimizing different rewards or goals at different levels of the hierarchy we share the rewards between the worker and the manager and introduce a weight hyper-parameter to regulate the effect of the rewards on each level.

% but similar to \cite{vezhnevets2017feudal} we share the rewards and introduce a weight hyper-parameter to regulate the effect of the rewards on each level.

% % \cite{bacon2017option} introduce a policy gradient learn options end-to-end but assume discrete options. 

% The lower level policy is referred to as the gating policy, policy over options, or manager while the higher level policy is refered to as the option policy, sub-policy, of worker. 

\section{Background}
%\subsection{Variational Hierarchical Recurrent Models for Dialog}
\label{sec:background_vhred}
A common approach to dialog modeling is to use a hierarchical seq2seq architecture, % \cite{ghandeharioun2019approximating,serban2017hierarchical}. 
such as the Variational Hierarchical Recurrent Encoder Decoder (VHRED) \cite{serban2017hierarchical}. We adopt VHRED here, following previous work which has found it to be the most effective version of several related architectures \cite{ghandeharioun2019approximating}. %jaques2019way

\begin{figure}[h]
\centering
\includegraphics[width=\linewidth]{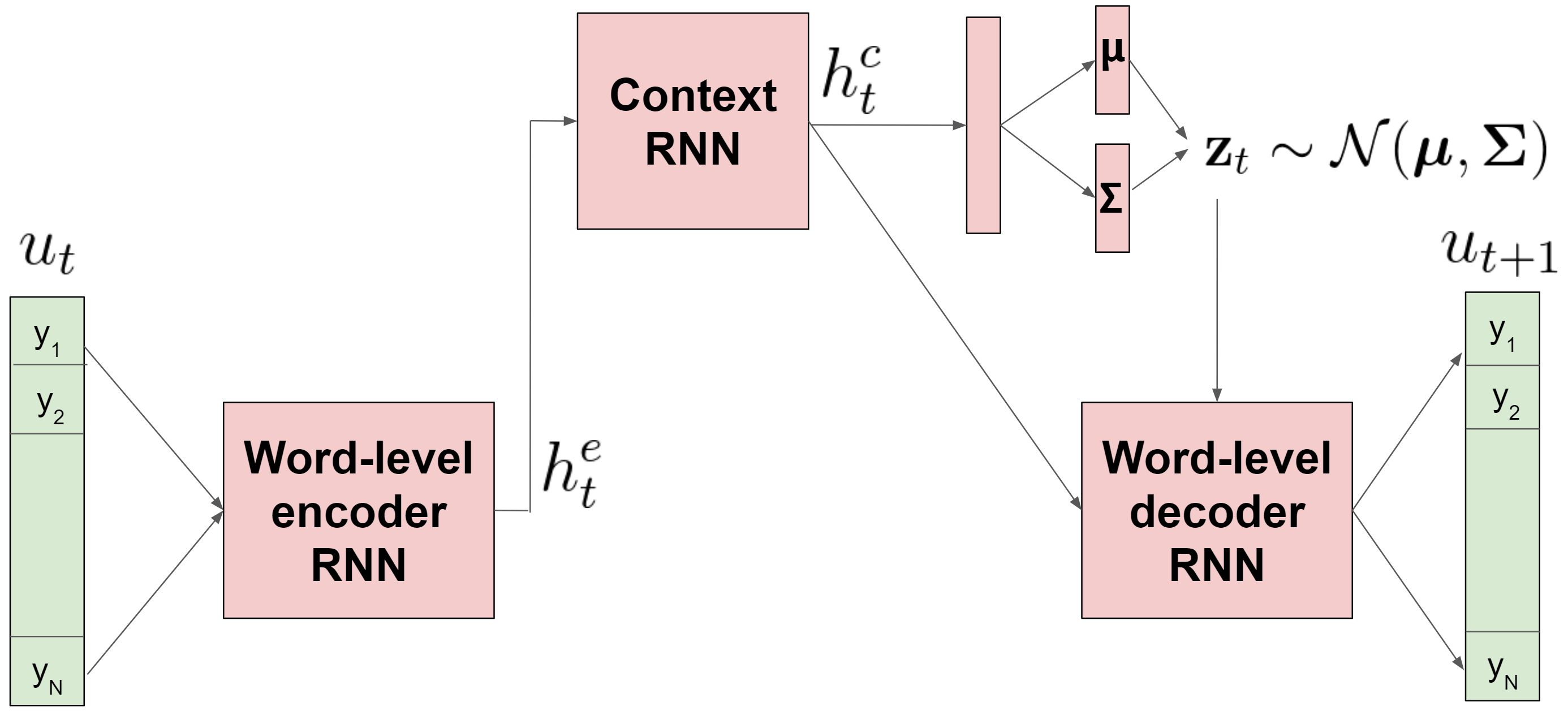}
\caption{VHRED model architecture, where the embedding vector $\mathbf{z}$ for each utterance is sampled from a multivariate normal distribution using the reparameterization trick.}
\label{fig:vhred}
\end{figure}

As shown in Figure \ref{fig:vhred}, VHRED uses three recurrent networks to generate the next utterance in a conversation. The word-level \textit{encoder RNN} operates on the words (tokens) of the input utterance $u_t = [y_1, y_2,... y_n]$, and encodes them into a representation $h^e_t = f^e(u_t)$. This is fed into a \textit{context RNN}, which forms the upper level of the hierarchy -- it is updated only after each utterance, rather than each token. Because it updates less frequently, the \textit{context RNN} is potentially better able to track longer-term aspects of the conversation. The \textit{context RNN} outputs $h^c_t = f^c(h^e_t)$, which is used to produce an utterance embedding $\mathbf{z}_t$. This is fed into the word-level \textit{decoder RNN} $f^d$, which produces the output utterance $u_{t+1}$, one token at a time. 

The model is similar to a variational autoencoder; $h^c_t$ is fed into fully connected layers that predict the mean $\boldsymbol{\mu}$ and variance $\boldsymbol{\Sigma}$ of a multivariate normal distribution. Through a KL-divergence constraint and the reparameterization trick, the model learns a probability distribution over the embedding vector $\mathbf{z}_t$ of each utterance, $p_{\boldsymbol{\theta}}(\mathbf{z}_t | u_{\leq t})$. Formally, the model can be described as follows:
\begin{align}
    h^e_{t} &= f^e(u_{t})  \label{eq:1} \\
    h^c_{t} &= f^c(u_{t}, h^e_{t})  \label{eq:2} \\
    \boldsymbol{\mu}, \boldsymbol{\Sigma} &= f(h^c_{t}) \\
    p_{\boldsymbol{\theta}}(\mathbf{z}_t | u_{\leq t}) &= \mathcal{N}(\mathbf{z}_t | \boldsymbol{\mu}, \boldsymbol{\Sigma}) \label{eq:4} \\
    p(u_{t+1} | u_{\leq t}) &= f^d(h^c_{t}, \mathbf{z}_t) \label{eq:5}
\end{align}
%Equations \eqref{eq:1}-\eqref{eq:5} 
%where $f^e$, $f^c$, and $f^d$ are Gated Recurrent Unit (GRU) networks for the encoder, context, and decoder RNNs, respectively.
%These additions are shown in purple in Figure \ref{fig:models}. 
%To further improve VHRED, we use knowledge distillation to increase the model's ability to recognize and encode the sentiment and semantics of the conversation, as proposed by \citet{ghandeharioun2019approximating}. 

\subsection{Reinforcement Learning}
% \natasha{Will write - Maybe refer to Deep Reinforcement Learning for
% Sequence-to-Sequence Models} 

We adopt the standard reinforcement learning framework where given the environment state $s \in \mathcal{S}$, an agent takes an action $a \in \mathcal{A}$ according to its policy $\pi: \mathcal{S} \times \mathcal{A} \rightarrow [0,1]$, and receives a reward $r: \mathcal{S} \times \mathcal{A} \rightarrow \mathbb{R}$. The environment then transitions to the next state according to the transition function $P: \mathcal{S} \times \mathcal{A} \times \mathcal{S} \rightarrow [0,1]$.
%acts in a Markov Decision Process consisting of a set of states $s \in \mathcal{S}$, actions $a \in \mathcal{A}$, state-transition probabilities $P: \mathcal{S} \times \mathcal{A} \rightarrow (S \rightarrow [0,1])$, and a reward function $r: \mathcal{S} \times \mathcal{A} \rightarrow \mathbb{R}$. A \textit{policy} $\pi: \mathcal{S} \times \mathcal{A} \rightarrow [0,1]$ is a mapping from states to the probabilities of selecting each action. 
The agent seeks to maximize the total expected future reward (long-term return):
\begin{equation}
    J(\pi) = \mathbb{E}_\pi \Big[ \sum_{t=0}^{\infty} \gamma^t \, r_{t+1} \, | \, s_0 \Big]
\end{equation}
given a starting state $s_0$ and a discount factor $\gamma \in [0,1]$.

\subsection{Policy Gradient Methods}
\label{sec:policy_grad}

% Policy gradient methods learn parameterized policies $\pi(a|s, \boldsymbol{\theta})$ for solving RL problems with $\boldsymbol{\theta} \in \mathbb{R}^{N_\theta}$ being a learned parameter vector. The policy gradient theorem \cite{sutton2000policy} derives the gradient of the expected return with respect to the policy parameters. In this paper, we use REINFORCE \cite{williams1992simple} which approximates the gradient at each time step $t$ by
% \begin{equation}
%     \nabla J(\pi_{\boldsymbol{\theta}}) \approx \nabla_\theta \ln \pi(a_t|s_t, \boldsymbol{\theta}) \sum_{k=t+1}^{T} \gamma^{k-t-1} r_{k}
% \end{equation}
% and uses gradient ascent to maximize the expected return for an interaction that ends at $T$. The above formulation is equivalent to minimizing the loss function,
% \begin{equation}
%     \mathcal{L}_\theta =  - \ln \pi(a_t|s_t, \boldsymbol{\theta}) \sum_{k=t+1}^{T} \gamma^{k-t-1} r_{k}
% \end{equation}

Policy gradient methods learn parameterized policies $\pi_{\boldsymbol{\theta}}(a|s)$ for solving RL problems with $\boldsymbol{\theta} \in \mathbb{R}^{N_\theta}$ being a learned parameter vector. The policy gradient theorem \cite{sutton2018reinforcement} derives the gradient of the expected return with respect to the policy parameters. In this paper, we use REINFORCE \cite{williams1992simple} which approximates the gradient at each time step $t$ using  
\begin{equation}
    \nabla J(\pi_{\boldsymbol{\theta}}) \approx R_t \, \nabla_{\boldsymbol{\theta}} \ln \pi_{\boldsymbol{\theta}}(a_t|s_t)
\end{equation}
where $R_t = \sum_{k=t+1}^{T} \gamma^{k-t-1} r_{k}$ is the observed future reward for an episode that ends at $T$. The expected return is maximized with gradient ascent. This is equivalent to minimizing the loss function $\mathcal{L}_{\boldsymbol{\theta}} =  - R_t \, \ln \pi_{\boldsymbol{\theta}}(a_t|s_t)$.

In continuous action spaces, actions $\mathbf{a} \in \mathbb{R}^{N_a}$ are sampled from a continuous probability distribution, such as a multivariate normal distribution. In this case, the policy $\pi$ can be parameterized as a probability density function over actions, 
\begin{multline} \label{eq:3}
    \pi_{\boldsymbol{\theta}}(\mathbf{a} | s) =\\[-4pt]
    \qquad  \frac{1}{\sqrt{(2\pi)^{N_a}|\boldsymbol\Sigma|}}
\exp\left(-\frac{1}{2}({\mathbf{a}}-\boldsymbol{\mu})^T{\boldsymbol{\Sigma}}^{-1}({\mathbf{a}}-\boldsymbol{\mu})
\right)
\end{multline}
\noindent where the actions are sampled from a multivariate normal  $\mathcal{N}\big(\boldsymbol{\mu}(\boldsymbol{s}; \boldsymbol{\theta}), \boldsymbol{\Sigma}(\boldsymbol{s}; \boldsymbol{\theta})\big)$. Here the mean $\boldsymbol{\mu}: \mathbb{R}^{N_s} \times \mathbb{R}^{N_\theta} \rightarrow \mathbb{R}^{N_a}$ and covariance matrix $\boldsymbol{\Sigma}: \mathbb{R}^{N_s} \times \mathbb{R}^{N_\theta} \rightarrow \mathbb{R}^{N_a \times N_a}$ are defined in terms of the current state $s$ and the policy parameters $\boldsymbol{\theta}$. The \textit{density} of the probability of actions, rather than the probability, is learned in the continuous case. We refer the reader to \citet{sutton2018reinforcement} for more details on extending policy gradient methods to the continuous case.

\section{Approach}
% \abdul{introduce approach}
% text text text
% \subsection{Variational Hierarchical Reinforcement Learning (VHRL) for Sequence Generation}
We pose dialog generation as an RL problem where the state, $s_t$, is all the previous dialog turns read by the model up to utterance $t$, and the rewards are calculated based on the dialog history and generated utterance. 

Previous approaches which have applied RL to language generation have done so at the word level, where the policy $\pi$ models the distribution over outputting the next word %only consider the output word distribution and treat it as the policy $\pi$ 
\cite{ranzato2015sequence,yu2017seqgan,li2016deep,li2017adversarial,jaques2019way}. %bahdanau2016actor
Instead, we cast our problem in the hierarchical reinforcement learning framework by considering the \textit{context RNN} as the \textit{manager} responsible for utterance-level decisions, and the \textit{decoder RNN} as the \textit{worker} responsible for word-level decisions. 

We leverage the fact that VHRED %(described in Section \ref{sec:background_vhred}), 
learns a probability distribution over latent variable $\mathbf{z}_t$ as a decision making component at the utterance level. Starting with a pre-trained VHRED model, we apply REINFORCE to tune the variational component, treating $\mathbf{z}_t$ as a continuous action. Thus, the manager policy is defined by the distribution of the prior latent variable $p_{\boldsymbol{\theta}}(\mathbf{z}_t|s_t)$, while the worker policy is the distribution of the output words $\pi_{\boldsymbol{\theta}}(\hat{y}_1, \dots, \hat{y}_{t}|\mathbf{z}_t, s_t)$, which is parameterized by the manager decisions. 

%\natasha{call back to figure and background section}

More specifically, the probability of a worker action $a_t$ is the joint probability of the generated utterance conditioned on the manager's decision $\mathbf{z}_t$,
\begin{equation}
\pi_{\boldsymbol{\theta}}(a_t|\mathbf{z}_t, s_t) = \prod_{t=1}^{T} \pi_{\boldsymbol{\theta}}(\hat{y}_t | \mathbf{z}_t, s_t,\hat{y}_1, \dots, \hat{y}_{t-1})
\end{equation}
while the probability of a manager action is given by the multivariate normal probability density function in Eq. \ref{eq:3}.

We propose a new approach which allows both the worker and manager to jointly optimize total expected future return by minimizing the following loss: 
\begin{equation}
    \mathcal{L}_{\boldsymbol{\theta}} = - \Big(\alpha R^m_t \ln p_{\boldsymbol{\theta}}(\mathbf{z}_t|s_t)
    + \beta R^w_t \ln \pi_{\boldsymbol{\theta}}(a_t|\mathbf{z}_t, s_t) \Big)
\end{equation}
where $R^m_t = \sum_{k=t+1}^{T} \gamma^{k-t-1} r^m_{k}$ is the manager's observed future reward and $R^w_t = \sum_{k=t+1}^{T} \gamma^{k-t-1} r^w_{k}$ is the worker's observed future reward. This formulation is analogous to REINFORCE as it shifts the model's decisions towards actions associated with positive rewards and discourages actions associated with negative rewards. The scalars $\alpha, \beta$ are hyperparameters used to regulate the effect of the rewards at each level of the hierarchy. %We start with a VHRED model pre-trained through maximum likelihood estimation and tune it with the proposed loss function to optimize for arbitrary conversation metrics. 
We call our approach Variational Hierarchical Reinforcement Learning (VHRL).

Unlike recently proposed HRL approaches which train the worker and manager separately as decoupled components \cite{kulkarni2016hierarchical,vezhnevets2017feudal,nachum2018data}, we train our entire model jointly, end-to-end. This implies that the worker (\textit{decoder RNN}) gradients flow through the manager (\textit{context RNN}), and both flow through the \textit{encoder RNN}. We make this decision for two reasons. First, $\mathbf{z}_t$ lives in a continuous high dimensional action space, making it difficult to learn a good policy $p_{\boldsymbol{\theta}}$ without a signal from the decoder. Second, this gives the decoder control over the representations learned by the encoder, facilitating optimization. As an ablation study, we experiment with decoupled decoder and encoder training, and find that the joint approach performs better.

The proposed loss allows for optimizing different rewards at different levels, which can be used to incorporate prior knowledge about the problem. For example, rewards relevant to the global dialog history could be considered only by the manager through $r^m_{k}$, rather than the worker. Conversely, rewards relevant to the word-by-word output could be considered by the worker through $r^w_k$ and not the manager. For simplicity, we optimize for all rewards at both levels (i.e. $r^w_{k} = r^m_{k}$) and achieve promising results.    

Similar to previous work applying RL to dialog \cite{xu2018towards,li2016deep} we use self-play to simulate the interactive environment in which the agent learns. We initialize conversations with randomly sampled starting sentences from the training set and let our model interact with a user simulator which is a fixed copy of itself. We limit each model to 3 turns for a total conversation length of 7 utterances. Limiting the length of simulated interactions is important since we found that long conversations are more likely to degenerate and go off-topic.

% \subsection{Self-Play for Simulating Interactions}

\section{Conversation Metrics} 
\label{sec:metrics}
Here we introduce several metrics for improving the quality of a conversation, which can be optimized using RL by treating them as rewards. Several metrics are inspired by previous work, but we also propose novel metrics such as toxicity. 

%\subsection{Sentiment}
\textbf{Sentiment: }
Emotion is important for creating a sense of understanding in human conversation \cite{weger2010active}. Building on previous work which used sentiment as a reward (e.g. \citet{shin2019happybot,jaques2019way}, we leverage a state-of-the-art sentiment detector, DeepMoji \cite{felbo2017using}, to reward generated utterances associated with positive sentiment emojis. The sentiment score is computed using the weights proposed by \citet{ghandeharioun2019approximating}.  %bodie2015role

%\subsection{Question}
\textbf{Question: }
Asking questions is an important active listening skill, and can improve the quality of interactions \cite{bodie2012listening}. Thus, we provide a positive reward when both a question word and a question mark are present in a generated response to encourage asking questions.

%\subsection{Repetition}
\textbf{Repetition: }
Repetitiveness has been frequently identified as a shortcoming of dialog models trained with MLE \cite{li2016deep}. We adopt a measure of repetitiveness recently proposed by \citet{see2019makes}, which was shown to be highly related to human judgments of conversation quality and engagement. Unlike previous work, we directly optimize for this metric using RL, rather than relying on conditional generation. To discourage repetition, our model receives a negative reward for repeating words it has produced in previous turns, excluding stop words and question words.

%\subsection{Semantic Similarity}
\textbf{Semantic Similarity: }
Paraphrasing and style matching are important in facilitating good conversation \cite{ireland2011language,weger2010active}, however most dialog models are not good at conditioning effectively on the conversation context \cite{sankar2019neural}. Therefore, we reward the cosine similarity between the simulated user and bot utterances in embedding space, as in \citet{see2019makes,jaques2019way}. However, instead of using word2vec embeddings we make use of the Universal Sentence Encoder \cite{cer2018universal} as it better correlates with human judgment when evaluating dialog quality \cite{dziri2019evaluating}.  

%\subsection{Toxicity}
\textbf{Toxicity: }
Open-domain dialog systems generate malicious, offensive, and biased language when trained on standard datasets scraped from online forums and movie scripts \cite{curry2018metoo,he2018detecting,henderson2018ethical}. We address this issue by penalizing our model for producing toxic responses as determined by a Naive Bayes-Logistic Regression classifier \cite{saleh2019team} trained on a dataset of 160K comments from the Toxic Comment Classification Challenge\footnote{\url{https://www.kaggle.com/c/jigsaw-toxic-comment-classification-challenge}}. The comments are labeled for toxicity, identity hate, obscenity, threats, and insults. We provide the probability of toxicity as a negative reward to penalize our dialog model for producing toxic responses.   

% notorious for producing offensive and aggressive language reflected in their training data. To discourage malicious outputs, we design a to 

\section{Experiments}
The goal of our experiments is to evaluate the effectiveness of VHRL for optimizing utterance level metrics like repetition and semantic similarity. We use both interactive human evaluation and automatic metrics. 

All of our models are trained on a corpus of 109K conversations scraped from \url{www.reddit.com/r/CasualConversations}, which was shown to result in higher conversation quality than traditional datasets such as Cornell movie dialogs \cite{ghandeharioun2019approximating}. We create two baselines by training on this dataset. The first is \textit{VHRED} \cite{serban2017deep}, described in Section \ref{sec:background_vhred}. The second is a \textit{Transformer} model \cite{vaswani2017attention} of comparable capacity for reference. We base our implementation of the Transformer on ParlAI \cite{miller2017parlai}.
%\abdul{Please add text explaining why we don't use BERT here}. 

% learning curves fig not cooperating
\begin{figure*}[ht]
\includegraphics[width=\linewidth]{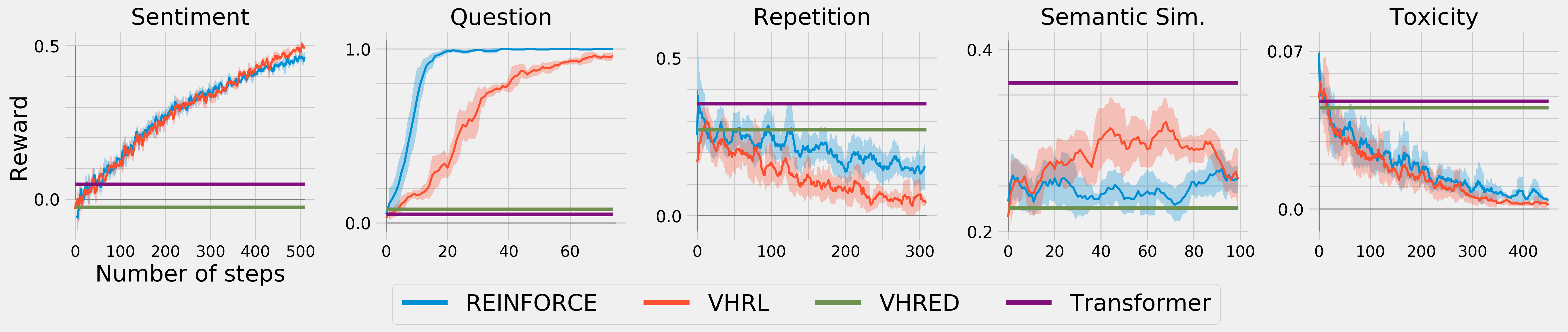}
\caption{Reward Learning curves for the proposed metrics. The $x$-axis represents number of RL training steps. The performance of the non-RL baselines is displayed for reference. REINFORCE and VHRL learn to outperform the baselines. Shaded area is standard deviation.}
\label{fig:learningcurves}
\end{figure*}

We test these dialog models against three RL techniques. We incorporate transfer learning by initializing all of our RL models with the pre-trained weights of the VHRED model. Our \textit{REINFORCE} model applies the REINFORCE algorithm described in Section \ref{sec:policy_grad} at the word-level, to affect the probability of generating each word in the output. In contrast, \textit{VHRL} incorporates additional rewards at the utterance-level to improve the continuous utterance embedding $\textbf{z}_t$. We compare these methods with a recently proposed approach for learning offline from a static batch of conversation data, \textit{Batch $\Psi$-learning} \cite{jaques2019way}. Finally, we include an ablated version of the VHRL model that uses decoupled training; \textit{i.e.} training alternates between optimizing the different levels of the hierarchy (manager and worker), with the crucial difference that the worker gradients are stopped so they do not update the manager. This \textit{Decoupled VHRL} ablation is more typical of standard HRL approaches used in maze and Atari games \cite{kulkarni2016hierarchical,vezhnevets2017feudal,nachum2018data}. All of our code is open-source at \url{https://github.com/natashamjaques/neuralchat}. Additional training details and model hyperparameters are given in the appendix. 

\subsection{Human Evaluation}
\label{sec:human_eval}
In addition to computing automatic measures, we conduct an interactive human evaluation, in which human users are able to chat with each of our bots in an open-ended, free-form setting. After chatting for at least three turns with each bot, users can end the conversation and rate the overall conversation quality, fluency, diversity, and the degree to which the bot's responses were contingent on the user's input. Because users can choose to chat as long as they like with any particular bot, we also measure chat length as a sign of engagement, following prior work \cite{zhou2018design}. %sidner2004look

We conduct two human evaluations by recruiting 100 Mechanical Turk workers to evaluate models on an open-source online platform at \url{https://neural.chat/} \cite{ghandeharioun2019approximating}. 
We are releasing all of the models for both studies publicly. Rather than cherry picking samples from the models, we encourage the reader to chat with each of the bots. The first study compares the quality of the proposed reward functions: \url{https://neural.chat/vhrl_rewards/}. The second study assesses the efficacy of the six proposed techniques when optimizing for all of the rewards jointly: \url{https://neural.chat/vhrl_techniques/}. We argue that this form of evaluation represents a more ambitious test of generalization than is typically attempted when deep RL algorithms are tested in the same environment in which they were trained, since human users are free to type any text they choose.

\section{Results and Discussion}

We first assess whether RL training can allow dialog agents to learn to optimize arbitrary, non-differential metrics of conversation quality. Table \ref{tab:samples} in Section \ref{sec:intro} shows samples of conversations from VHRL trained with each of the rewards, and Figure \ref{fig:learningcurves} shows the performance of the RL and baseline models on those five metrics. The RL models are able to dramatically improve generated conversations above the baseline VHRED model with which they are initialized, improving sentiment and semantic similarity, asking more questions, and reducing repetition and toxicity. The RL models also improve significantly over the performance of the Transformer model on these metrics, with the exception of the semantic similarity metric. We note that compared to the VHRED architecture, the Transformer has higher similarity but is also much more repetitive.  

While both REINFORCE and VHRL are equally able to learn to optimize toxicity and sentiment, VHRL out-performs REINFORCE on repetition and semantic similarity. We believe this is because sentiment and toxicity are more closely linked to the choice of words used to form a sentence, and thus are able to be learned at the word-level. In contrast, modeling whether a sentence has occurred earlier in the conversation and is thus being repeated is much harder to learn at word-level granularity, and can be optimized more easily at the utterance-level using VHRL. Similarly, making a response more similar to the previous response is also better modeled at the utterance-level. 

Note that REINFORCE outperforms VHRL on the question metric. This is because the model quickly learns to ask a single, repeated question, allowing it to trivially optimize the reward function. %(the repetitiveness of the REINFORCE question model is \abdul{\#}, while the VHRL question model is \abdul{\#}). %REINFORCE has a known tendency to converge quickly to suboptimal functions \cite{duan2016benchmarking}. 
Using reward functions which are too easily exploitable can limit the effectiveness of RL for dialog, a finding also noted in \citet{jaques2019way}. Here we propose new reward functions, such as toxicity, that are less easy to exploit. By optimizing a combination of these rewards with sometimes conflicting objectives (as we explore in Section \ref{sec:results_combined}), we can show that the reward function is difficult to trivially exploit, as suggested by \citet{deb2014multi}.

\begin{figure}[ht]
\centering
\includegraphics[width=.7\linewidth]{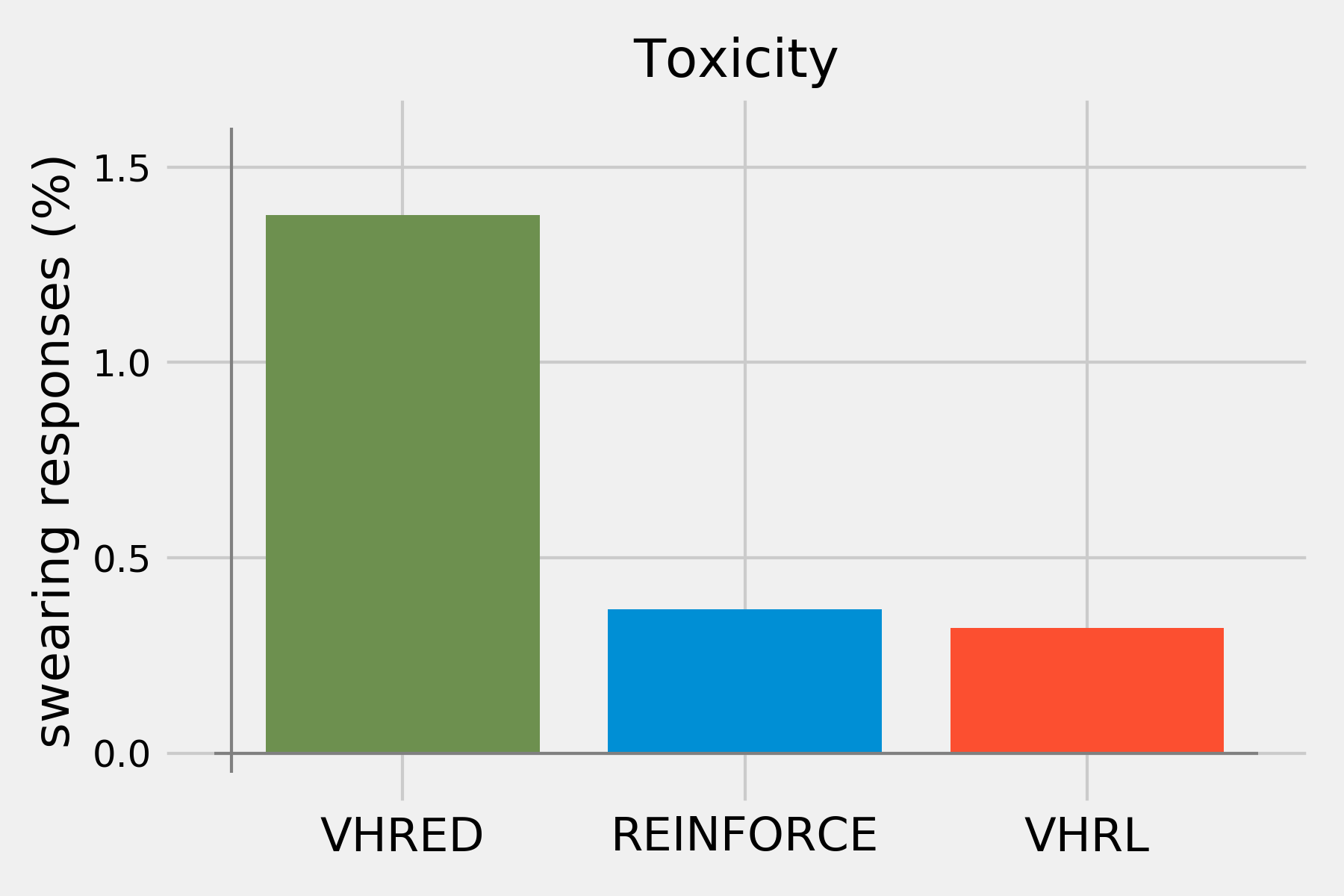}
\caption{Training with RL to reduce toxicity decreases the percentage of generated utterances containing swear words.}
\label{fig:toxicitybarplot}
\end{figure}

As an additional post-hoc test of whether reducing our toxicity metric actually reduces undesirable behaviors in dialog models, we count the number of swear words used by each model in response to the 10,000 utterances in the test set. Figure \ref{fig:toxicitybarplot} shows the results. The baseline VHRED model uses a swear word in 1.5\% of the responses, while using RL to lower the toxicity reduces swearing to less than one third of the original amount. 

We conducted an interactive human evaluation, as described in Section \ref{sec:human_eval}, in order to assess how well the proposed reward functions relate to human judgments of conversation quality when optimized with REINFORCE or VHRL; the results are presented in Table \ref{tab:rewards}. % shows human ratings of the conversations in terms of the overall quality, fluency, diversity, and contingency, the total score for each bot (sum of all the ratings), and the total conversation length in number of turns. 
Each bot was trained with respect to one of the 5 reward functions, and the results are ordered from least to highest scoring rewards in terms of summed human ratings. As is evident in the table, the VHRL model trained to optimize for asking questions achieved the highest ratings, followed by VHRL minimizing repetition, and VHRL minimizing toxicity. This provides evidence that our proposed rewards lead to enhanced conversation quality as judged by human users, and that VHRL provides the most effective method for learning them. %While VHRL did not significantly outperform REINFORCE for the sentiment and similarity rewards, we found that the similarity reward performed worst overall in terms of human judgment. This mirrors a finding in previous work, which showed that similarity in GloVe embedding space between input and response was not strongly correlated with human judgments of conversation quality \cite{see2019makes}. While here we propose using USE similarity instead, and find that our models are able to optimize this metric (see Figure \ref{fig:learningcurves}), models trained to optimize solely for USE similarity produce conversations of lower quality than those trained with other metrics. 

% Techniques table - ordering based on score
\begin{table*}[ht]
\centering
% \ra{1.0}
\noindent\makebox[\textwidth]{
\addtolength{\tabcolsep}{0pt}
\begin{tabular}{lccccc|cc}\toprule
Model& Quality & Fluency & Diversity & Contingency & Total & Chat Len.\\ \midrule
Batch $\Psi$ \cite{jaques2019way} & 2.17 & 3.89 & 3.13 & 1.98  & 11.17 & 11.44 \\
Decoupled VHRL (ablation) & 2.46 & 4.15 & 3.61 & 2.02  & 12.24 & 12.14 \\
Transformer & 2.62 & 4.17 & 3.23 & 2.34  & 12.36 & 11.28 \\
REINFORCE & 2.89 & 4.47 & 3.67 & \textbf{2.80}  & 13.84 & 11.60\\
VHRED & 2.84 & 4.53 & \textbf{4.43} & 2.47  & 14.27 & 10.94\\
VHRL (ours) & \textbf{2.91} & \textbf{4.65} & 4.26 & 2.67  & \textbf{14.49} & \textbf{12.84}\\
\bottomrule
\end{tabular}}
\caption{Interactive human evaluation results comparing different RL training approaches optimizing for all five rewards, ordered by overall total rating score. Ratings are on a Likert scale (1-7).}
\label{tab:techniques}
\end{table*}

\begin{table}[ht]
% \ra{0.9}
\small\addtolength{\tabcolsep}{-4.3pt}
\begin{tabular}{rccccc|c}\toprule
\multicolumn{1}{l}{Model} & Quality & Fluency & Diversity & Cont. & Total & Chat Len\\ \midrule

\multicolumn{1}{l}{\textbf{Similarity}}&&&&&\\
REINFORCE\phantom{.} & \underline{2.71} & \underline{4.20} & \underline{3.86} & \underline{2.73} & \underline{12.10} & \underline{12.36}\\
VHRL & 2.51 & 3.92 & 3.67 & 2.22  & 11.14 & 11.56\\

\multicolumn{1}{l}{\textbf{Sentiment}}&&&&& \\
REINFORCE\phantom{.} & \underline{2.80} & \underline{4.55} & 3.90 & 2.43 & \underline{12.43} & 11.02\\
VHRL & 2.72 & 4.30 & \underline{4.32} & \underline{2.50} & 12.28 & \underline{11.12}\\

\multicolumn{1}{l}{\textbf{Toxicity}}&&&&&\\
REINFORCE\phantom{.} & 2.71 & 4.12 & 4.06 & 2.55  & 11.98 & 10.74\\
VHRL & \underline{2.76} & \underline{4.58} & \underline{4.34} & \underline{2.64} & \underline{12.82} & \underline{11.32}\\

\multicolumn{1}{l}{\textbf{Repetition}}&&&&&\\
REINFORCE\phantom{.} & 2.74 & 4.02 & 4.28 & 2.30 &  11.92 & \underline{11.52}\\
VHRL & \underline{3.00} & \underline{4.39} & \underline{4.41} & \underline{2.84} & \underline{13.12} & 10.66\\

\multicolumn{1}{l}{\textbf{Question}}&&&&&\\
REINFORCE\phantom{.} & 2.39 & 4.08 & 2.45 & 2.31 & 9.80 & \underline{\textbf{12.98}}\\
VHRL & \underline{\textbf{3.27}} & \underline{\textbf{4.86}} & \underline{\textbf{4.47}}& \underline{\textbf{2.88}}  & \underline{\textbf{14.14}} & 11.68\\

\bottomrule
\end{tabular}
\caption{Interactive human evaluation results comparing the proposed reward functions, REINFORCE, and VHRL. Ratings are on Likert scale (1-7). Higher is better.}
\label{tab:rewards}
\end{table}

%\subsection{Human evaluation}

%\subsection{Automatic Evaluation}

\subsection{Comparison of RL techniques for learning combined reward}
\label{sec:results_combined}
As described in the previous section, optimizing for an overly simplistic metric (such as asking questions) can lead algorithms such as REINFORCE to trivially exploit the reward function at the expense of conversation quality. The five metrics proposed here do not fully encompass what it means to have a good conversation with a human user when optimized separately. Previous work found that optimizing individual metrics can actually reduce human judgments of conversation quality below the score of the MLE baseline \cite{jaques2019way}. 

Therefore, instead of optimizing for individual metrics, we also train a variety of models to optimize for a combination of all five proposed rewards\footnote{The weights placed on each reward are given in the appendix}, making the reward function more complex and less easily exploited. The results are shown in Table \ref{tab:techniques}, which is ordered from least to highest summed human ratings. All models proposed here, including the MLE baselines, outperform prior work by \citet{jaques2019way}. The ablated version of our approach, \textit{Decoupled VHRL}, performs poorly, suggesting our proposed joint training scheme for VHRL is an important component of the algorithm. 
% Techniques table - ordering based on model
% \begin{table*}[h]
% \centering
% \ra{1.0}
% \noindent\makebox[\textwidth]{
% \addtolength{\tabcolsep}{0pt}
% \begin{tabular}{lccccc|cc}\toprule
% Model& Quality & Fluency & Diversity & Contingency & Total & Chat Len.\\ \midrule

% Transformer & 2.62 & 4.17 & 3.23 & 2.34  & 12.36 & 5.64 \\
% VHRED & 2.84 & 4.53 & 4.43 & 2.47  & 14.27 & 5.47\\
% Batch $\Psi$ \cite{jaques2019way} & 2.17 & 3.89 & 3.13 & 1.98  & 11.17 & 5.72 \\
% REINFORCE & 2.89 & 4.47 & 3.67 & 2.80  & 13.84 & 5.80\\
% Alternate VHRL & 2.46 & 4.15 & 3.61 & 2.02  & 12.24 & 6.07 \\
% VHRL (ours) & 2.91 & 4.65 & 4.26 & 2.67  & 14.49 & 6.42\\
% \bottomrule
% \end{tabular}}
% \caption{Interactive human evaluation results comparing different RL training approaches optimizing for multiple rewards. Ratings are on Likert scale. No inapprop}
% \end{table*}

VHRED, REINFORCE, and VHRL all exceed the performance of the Transformer in terms of human judgments of conversation quality. While recent advances in language modeling with Transformer architectures (e.g. \citet{radford2019language}) are quite promising, translating these successes into conversational AI is still an ongoing area of research.
%Transformers have been shown to be invariant to perturbing the order of words in the input \cite{sankar2019neural}, which could possibly be a hindrance for dyadic dialog which requires responding in a way that is contingent on the partner's utterance. 
Models such as VHRED have an architecture designed for the dyadic nature of the conversation (with an explicit update based on each conversation turn); in contrast, the Transformer has no special architectural features to denote conversation turns beyond the \textit{$<$end of utterance$>$} token. Recurrent models have been shown to be more adept at capturing and exploiting hierarchical information in language \cite{tran2018importance}. Further, Transformers have been shown to be less sensitive to the conversation history relative to recurrent models \cite{sankar2019neural}. We have observed that the Transformer is highly repetitive, and vulnerable to monologuing rather than generating answers contingent on the user's input. Although it has a high score on the semantic similarity metric in Figure \ref{fig:learningcurves}, Table \ref{tab:techniques} demonstrates that this does not translate into improved contingency ratings. 
%a form of monologuing behavior, where when the user gives short responses, the Transformer appears to base its next conversation turn by mainly conditioning on its own previous turn. 
%\natasha{Do we have evidence to back this up? A sample conversation, or some attention weights?} 
%\abdul{Might not be exactly what you are looking for but this paper shows that recurrent models are better at capturing hierarchical structure which extends to dialog obviously 'The Importance of Being Recurrent for Modeling Hierarchical Structure'} \cite{sankar2019neural}
%\abdul{Another thing is that figure 1 in 'Do neural dialog systems use history effectively' shows that recurrent models consider more of the history relative to transformers which could explain repetitiveness}

Finally, in comparing the RL techniques to the VHRED baseline on which they were based, we see that a na{\"i}ve application of the REINFORCE algorithm does not lead to overall improvements in human ratings of conversation quality. While the language generated by the REINFORCE model is less toxic and more positive, this comes at the cost of a slight reduction in overall conversation quality.   %This is most likely because REINFORCE is known to converge quickly to suboptimal policies \cite{duan2016benchmarking}, and may be able to trivially exploit the reward functions at the cost of truly engaging and interesting conversation. 
In contrast, VHRL is the only technique that allows the model to optimize for reducing toxicity, improving sentiment, etc., while increasing the overall human judgments of the quality of the conversation. Note that the chat length is higher with VHRL, suggesting users are more interested and engaged when chatting with VHRL versus the other models. Thus, VHRL can be used to optimize for metrics that make the dialog model more safe and appropriate for a particular application domain, while maintaining the ability to have an enjoyable and engaging conversation with human users.

\section{Conclusion}
We have demonstrated that RL can be used to improve the outputs of an open-domain dialog model with respect to human-centered metrics of conversation quality. For example, RL can reduce the toxicity of the generated language, a problem that has previously hindered deployment of these systems to the real world. By developing metrics tailored to a particular application domain (for example, increasing politeness for a technical-support system), these techniques could be used to help open-domain dialog models integrate with real-world products. We have shown that our proposed VHRL technique is most effective for optimizing long-term conversation rewards, and for improving conversation quality while improving metrics like toxicity.

% Lowered font size for bibliography to minimum allowed 
\bibliographystyle{aaai}
\fontsize{9.0pt}{10.0pt}\selectfont
\bibliography{dialog}

\newpage

% \vspace{5cm}
% \usepackage[up,bf,raggedright]{titlesec}

\appendix
\onecolumn

\begin{changemargin}{1.6cm}{1.6cm}

\section*{\huge{Supplemental Material \hfill}}

\vspace{0.5cm}
\section{Additional Chat Examples \hfill}
\label{sec:chatexamples}
Refer to table \ref{tab:extendedsamples} for extended chat examples of VHRL tuned to optimize for different rewards.
\begin{table*}
\ra{0.6}
\noindent\makebox[\textwidth]{
%\begin{tabular}{p{2.6cm}|p{2.6cm}|p{2.6cm}|p{2.6cm}|p{2.6cm}|p{2.3cm}}
\begin{tabular}{p{1.8cm}|p{8.7cm}}
\normalsize Model & Chat Example \\ 
\midrule
Transformer &
\small
\makecell[lp{8.6cm}]{[Usr]: hey! do you like music?
\\ {[}Bot{]}: what kind of music?
\\ {[}Usr{]}: any kind. what do you listen to usually?
\\ {[}Bot{]}: what kind of stuff do you listen to?
\\ {[}Usr{]}: i like pop. what about you?
\\ {[}Bot{]}: what kind of stuff do you listen to?}\\ 
\midrule
VHRED &
\small
\makecell[lp{8.6cm}]{[Usr]: hey
\\ {[}Bot{]}: hi. nice to meet you.
\\ {[}Usr{]}: how are you today?
\\ {[}Bot{]}: i'm currently in my third year of high school and i've been looking for a job for about a year now and i'm really excited
\\ {[}Usr{]}: ohh cool good luck looking for jobs
\\ {[}Bot{]}: what is the most dangerous thing you've ever done?
\\ {[}Usr{]}: i once road my bike without a helmet!}\\ 
\midrule
\makecell[lp{1.8cm}]{VHRL\\ Sentiment} &
\small
\makecell[lp{8.6cm}]{[Usr]: hello!
\\ {[}Bot{]}: i'm in my sunroom a lot and it's a great time!
\\ {[}Usr{]}: that sounds fun. how are you?
\\ {[}Bot{]}: it's a lot of fun, and it's great that you've got a lot of hobbies, what are you interested in now?
\\ {[}Usr{]}: i like reading books and riding my bike around boston
\\ {[}Bot{]}: i'm so excited for you!
\\ {[}Usr{]}: thanks! i appreciate that
\\ {[}Bot{]}: you're a good person}\\ 
\midrule
VHRL Question &
\small
\makecell[lp{8.6cm}]{[Usr]: hi there!
\\ {[}Bot{]}: what do you do for fun?
\\ {[}Usr{]}: i like reading books.
\\ {[}Bot{]}: what do you like to do in your life?
\\ {[}Usr{]}: i like hanging our with friends and taking fun classes at school.
\\ {[}Bot{]}: i'm in the same boat.
\\ {[}Usr{]}: ohh cool.
\\ {[}Bot{]}: like, what kind of music do you like to listen to in general?
\\ {[}Usr{]}: i like listening to pop}\\ 
\midrule
VHRL Semantic Similarity &
\small
\makecell[lp{8.6cm}]{[Usr]: do you have any hobbies?
\\ {[}Bot{]}: sure.
\\ {[}Usr{]}: what are your hobbies then?
\\ {[}Bot{]}: i like to play video games too.
\\ {[}Usr{]}: ohh cool i also like video games. do you have a favorite game?
\\ {[}Bot{]}: do you play?
\\ {[}Usr{]}: yeah i play battlefield sometimes.}\\ 
\midrule
\makecell[lp{1.8cm}]{VHRL\\ Toxicity} &
\small
\makecell[lp{8.7cm}]{[Usr]: you're an idiot
\\ {[}Bot{]}: i'm in my sunroom a lot and it's a big deal to me.
\\ {[}Usr{]}: that's just stupid.
\\ {[}Bot{]}: i'm sorry you had to go through that.
\\ {[}Usr{]}: i hate you
\\ {[}Bot{]}: i'm not sure if this is the right word but i'm not sure what to do}\\

\bottomrule

\end{tabular}}
\caption{Chat examples showing that our proposed VHRL technique can be used to optimize for human-centred rewards.}
\label{tab:extendedsamples}
\end{table*}

\section{Training Details \hfill}
\label{sec:hparam}
The hyperparameters for the baseline VHRED model we trained were as follows: encoder RNN hidden size = 1250, context RNN hidden size = 1000, decoder RNN hidden size = 1250, $\mathbf{z}$ embedding size = 600, gradient clip = 1.0, dropout d = 0.2. We used one GRU layer for each of the word encoder, context RNN, and decoder. \\ \\
\noindent We also trained the ParlAI \cite{miller2017parlai} implementation of the Transformer \cite{vaswani2017attention} of comparable size to the VHRED model. The hyperparametrs were: encoder layers = 8, decoder layers = 8, attention heads = 8, embedding size = 512, and feedforward layer size = 2048. \\ \\
For all our experiments, the vocabulary size was limited to the 20K most common words. The maximum conversation length was fixed at 5 utterances (context from more than 5 utterances ago was discarded), and the maximum sentence length was 30 tokens.\\ \\
For the RL models, all the rewards were standardized by subtracting the mean and dividing by the standard deviation. The combined reward signal provided to the models took the form of 
\small{\begin{align*}
    0.15 * \texttt{sentiment} + 0.25 * \texttt{question} + 0.5 * \texttt{repetition} + 0.05 *  \texttt{similarity} + 0.05 * \texttt{toxicity}
\end{align*}}
\normalsize
\section{Human Evaluation Interface \hfill}
\label{sec:interface}
The models used for human evaluation were deployed online at \url{https://neural.chat/}. Figures \ref{fig:chat} and \ref{fig:evaluation} show the chat and evaluation interfaces. The Amazon Mechanical Turk workers were redirected to the study links \url{https://neural.chat/vhrl_rewards/} and \url{https://neural.chat/vhrl_techniques/} after which they were presented with the models in random order. The MTurk workers rated each model before moving on to the next one. 

\newpage
\begin{figure}[H]
\centering
\includegraphics[width=0.8\linewidth]{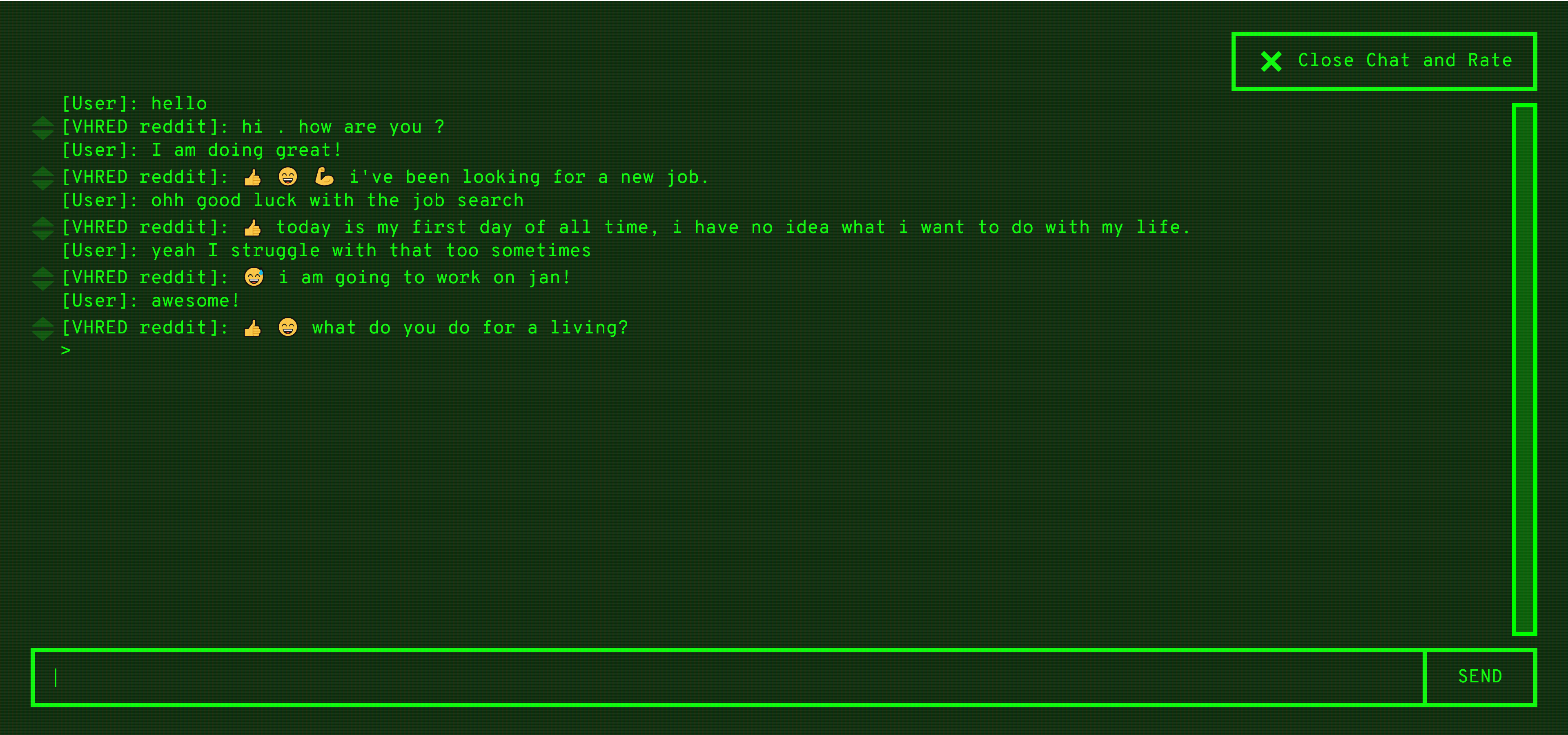}
\caption{Interactive human evaluation chat interface available at \url{https://neural.chat/}}
\label{fig:chat}
\end{figure}

\begin{figure}[H]
\centering
\includegraphics[width=0.8\linewidth]{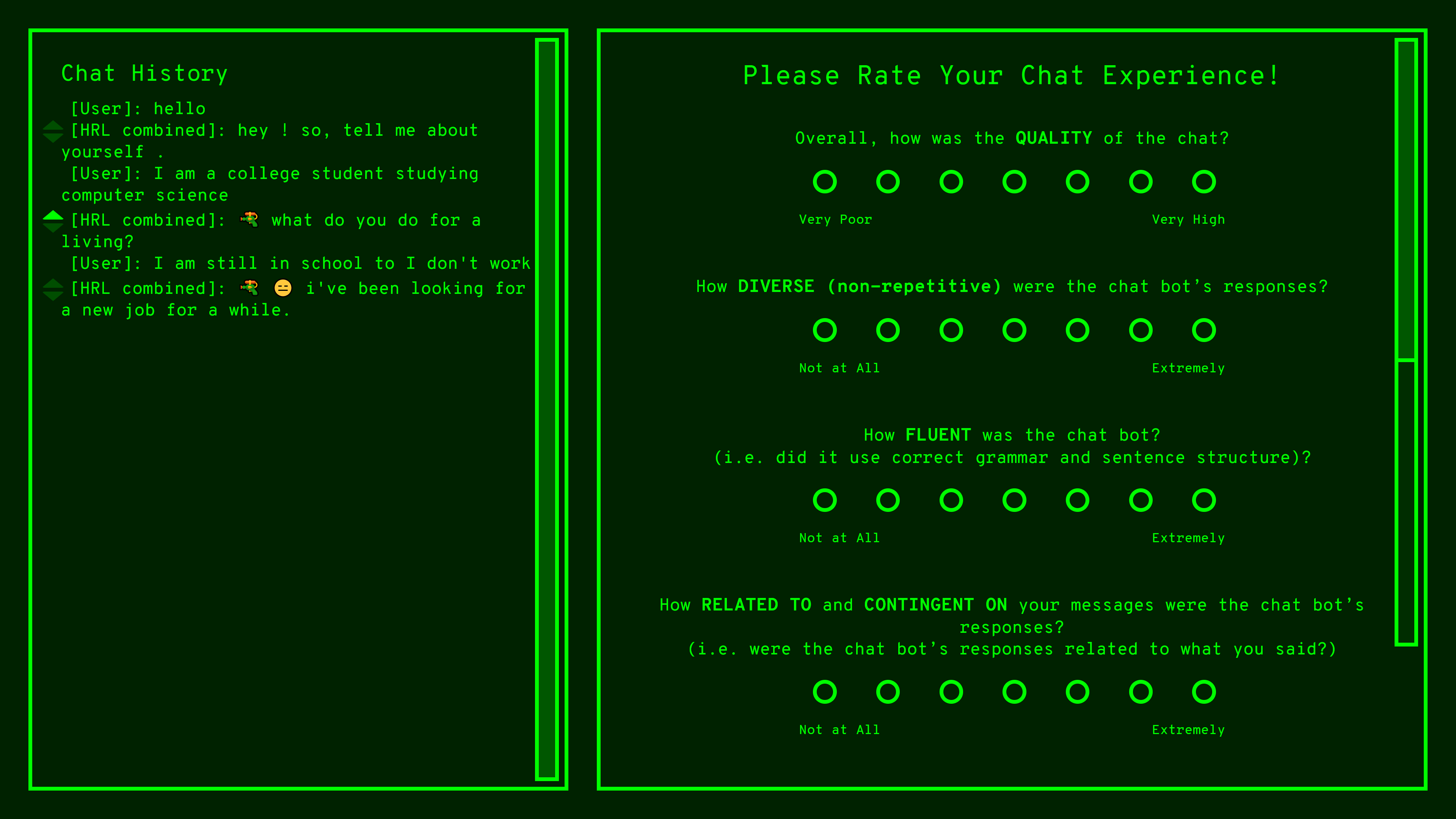}
\caption{Ratings page for human evaluations available at \url{https://neural.chat/}}
\label{fig:evaluation}
\end{figure}

\end{changemargin}

\end{document}